\documentclass{article}

\usepackage{microtype}
\usepackage{graphicx}
\usepackage{subfigure}
\usepackage{booktabs} 
\usepackage{CJKutf8}
\usepackage{amsmath}

\usepackage{hyperref}

\usepackage[accepted]{icml2021}

\icmltitlerunning{Presidifussion: Few-shot President’s Calligraphy Style Learning}

\begin{document}

\twocolumn[
\icmltitle{Few-shot Calligraphy Style Learning}




\begin{icmlauthorlist}
\icmlauthor{Fangda Chen}{1}
\icmlauthor{Jiacheng Nie}{2}
\icmlauthor{Lichuan Jiang}{3}
\icmlauthor{Zhuoer Zeng}{4}
\end{icmlauthorlist}

\icmlaffiliation{1}{121090029}
\icmlaffiliation{2}{121090422}
\icmlaffiliation{3}{121090225}
\icmlaffiliation{4}{120040045}

\icmlkeywords{Machine Learning, ICML}

\vskip 0.3in
]




\begin{abstract}
We introduced "Presidifussion," a novel approach to learning and replicating the unique style of calligraphy of President Xu, using a pretrained diffusion model adapted through a two-stage training process. Initially, our model is pretrained on a diverse dataset containing works from various calligraphers. This is followed by fine-tuning on a smaller, specialized dataset of President Xu's calligraphy, comprising just under 200 images. Our method introduces innovative techniques of font image conditioning and stroke information conditioning, enabling the model to capture the intricate structural elements of Chinese characters. The effectiveness of our approach is demonstrated through a comparison with traditional methods like zi2zi and CalliGAN, with our model achieving comparable performance using significantly smaller datasets and reduced computational resources. This work not only presents a breakthrough in the digital preservation of calligraphic art but also sets a new standard for data-efficient generative modeling in the domain of cultural heritage digitization.

\end{abstract}

\section{Introduction}

In the realm of Chinese calligraphy, where each stroke and character is imbued with centuries of artistic tradition and cultural significance, the challenge of digital preservation and replication is particularly pronounced. The intersection of culture and computational technology has sparked interest in this area, with a focus on using generative models, such as Generative Adversarial Networks (GANs) like CalliGAN \cite{wu2020calligan}, to replicate calligraphic styles. However, these models typically require extensive datasets to effectively learn and reproduce the complexities of calligraphy. This poses a significant challenge, especially in cases like President Xu's calligraphy, which, despite being prevalent across the CUHK-Shenzhen campus, still does not provide access to the thousands of samples usually needed for GANs.

\begin{figure}[ht]
\vskip 0.2in
\begin{center}
\centerline{\includegraphics[width=\columnwidth]{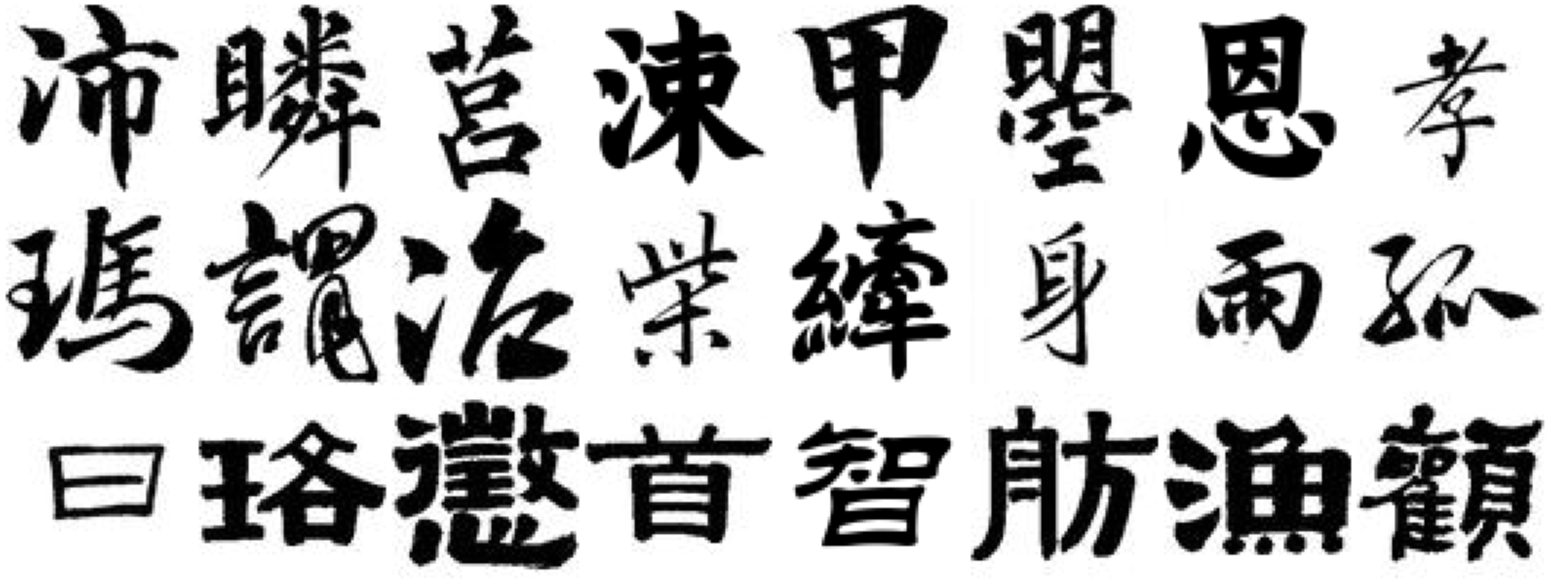}}
\caption{Samples of ancient artworks of calligraphy. The three rows of characters stand for the style of Regular Script, Semi-cursive Script, and Clerical Script, respectively.}
\label{sample}
\end{center}
\vskip -0.2in
\end{figure}

To address this challenge, we propose a novel two-stage approach within the framework of diffusion models. We begin by pretraining on a comprehensive dataset comprising works from a range of ancient calligraphers shown in figure~\ref{sample}, laying the groundwork for understanding the broader strokes of the art. Subsequently, we refine our model's understanding by fine-tuning on a much smaller, curated dataset from President Xu, achieving an intimate grasp of the subtle nuances that define his style. This process culminates in a model adept at generating calligraphy that not only embodies the general aesthetics of Chinese characters but also resonates with the distinct flair of President Xu's brushwork.

In this paper, we present a few-shot finetuning method that requires only a small subset of the target calligrapher’s writings. Our approach necessitates significantly fewer computational resources than traditional methods and still delivers satisfying outcomes. To augment the capability of our diffusion model, we introduce font image conditioning and stroke information conditioning. These innovative conditioning techniques inject precise structural information into the diffusion process, thereby enhancing the structural correctness of the generated characters—a metric we quantify using the structural similarity index (SSIM).

While we have made strides in the structural replication of calligraphy, our method is not without limitations, and the pursuit of perfection in digital calligraphy generation demands further exploration. Thus, our paper not only showcases our current achievements but also lays the groundwork for future advancements in the field.

\section{Method}
We employ a diffusion model and introduce a two-stage approach to train a generative model for calligraphy. Initially, we pretrain the model on a large dataset composed of works from ancient, renowned calligraphers. Subsequently, we fine-tune the pretrained model on a smaller, targeted dataset from a specific calligrapher. This methodology results in a model capable of generating new works that mimic the style of the chosen calligrapher.
\subsection{Preliminary: Denoising Diffusion Models}

Denoising diffusion models\cite{ho2020denoising} are a class of generative models that transform a simple distribution, typically Gaussian noise, into a complex data distribution through a Markov chain. The process involves two key phases: the forward diffusion process and the reverse generative process. In the forward process, data is gradually corrupted by adding noise over a sequence of time steps, typically modeled by $x_{t} = \sqrt{\alpha_{t}}x_{t-1} + \sqrt{1 - \alpha_{t}}\epsilon$, where $\epsilon \sim \mathcal{N}(0, I)$ and $\{\alpha_{t}\}_{t=1}^{T}$ are variance schedules. The reverse process aims to learn to denoise, effectively reversing the diffusion by modeling the distribution $p_{\theta}(x_{t-1}|x_{t})$. Training is conducted by minimizing a variant of the variational lower bound, often resembling a denoising score matching objective, which encourages the model to reconstruct the data from its noised version. These models have shown remarkable performance in generating high-quality samples in various domains.

We followed the textbook Understand Deep Learning \cite{prince2023understanding} to use a model $\hat\epsilon = \boldsymbol{g}(\boldsymbol{z}_t, \phi, t)$ to predict the noise. The final loss function for a dataset of size $I$ would be
$$
L(\phi)=\sum_{i=1}^I\sum_{t=1}^T||\boldsymbol{g}(\sqrt{\alpha_t}\cdot\boldsymbol{x}_i+\sqrt{1-\alpha_t}\cdot\epsilon_{it}, \phi, t)-\epsilon_{it}||^2
$$

\subsection{Model Architecture}
\begin{figure}[ht]
\vskip 0.2in
\begin{center}
\centerline{\includegraphics[width=\columnwidth]{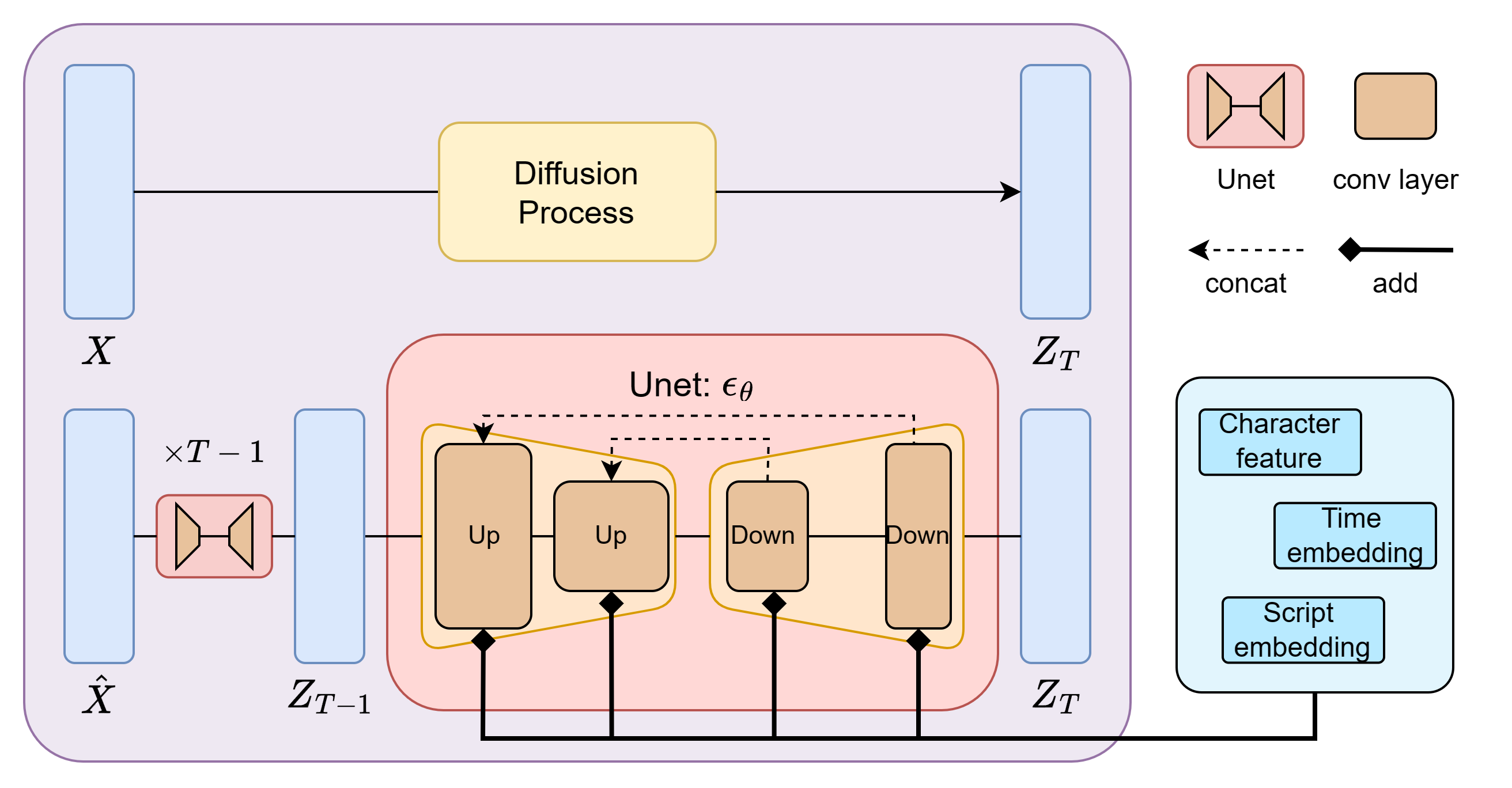}}
\caption{Model architecture}
\label{ModelArch}
\end{center}
\vskip -0.2in
\end{figure}

To facilitate calligraphy diffusion, we have incorporated denoising diffusion as the fundamental component of our methodology, as shown in figure~\ref{ModelArch}. Each character is treated as an image and a DDPM model is employed to learn the distribution of such images. In each denoising step, a U-net is used to predict the noise.

In order to precisely generate target characters, we leverage condition embedding empowered by font and utilize transformer for controlling the diffusion process. To train the conditional denoising diffusion for xys-style calligraphy with a limited dataset of less than 500 samples, we initially trained a general CDDPM on a significantly larger dataset encompassing diverse calligraphic styles from various individuals. Subsequently, fine-tuning was performed using the xys-dataset to refine the model's performance.

\subsection{Conditional Diffusion}
We give the model conditional guidance by adding the condition embedding ad an input to the model, i.e, $\hat\epsilon = \boldsymbol{g}(\boldsymbol{z}_t, \phi, t, \boldsymbol{c})$, where $\boldsymbol{c}$ is the condition embedding.

A conventional approach to generate condition embedding involves encoding each character in the dictionary used for the diffusion model and incorporating this coding into every down-sample or up-sample step of the Unet, similar to how Unet's time embedding is utilized. However, the quality of generated images using this type of Condition embedding is not satisfactory. Hence, we propose two distinct methods for condition embedding to enhance classification accuracy and provide more comprehensive information regarding character distribution for our model.

Beside the two conditioning methods, we also conduct experiments on the naive way that randomly initializes the embedding vector from a standard Guassian distribution as a comparision group. We simply implement this using nn.Embedding in PyTorch.

\subsubsection{Font Image Conditioning}

\begin{figure}[ht]
\vskip 0.2in
\begin{center}
\centerline{\includegraphics[width=1.0\linewidth]{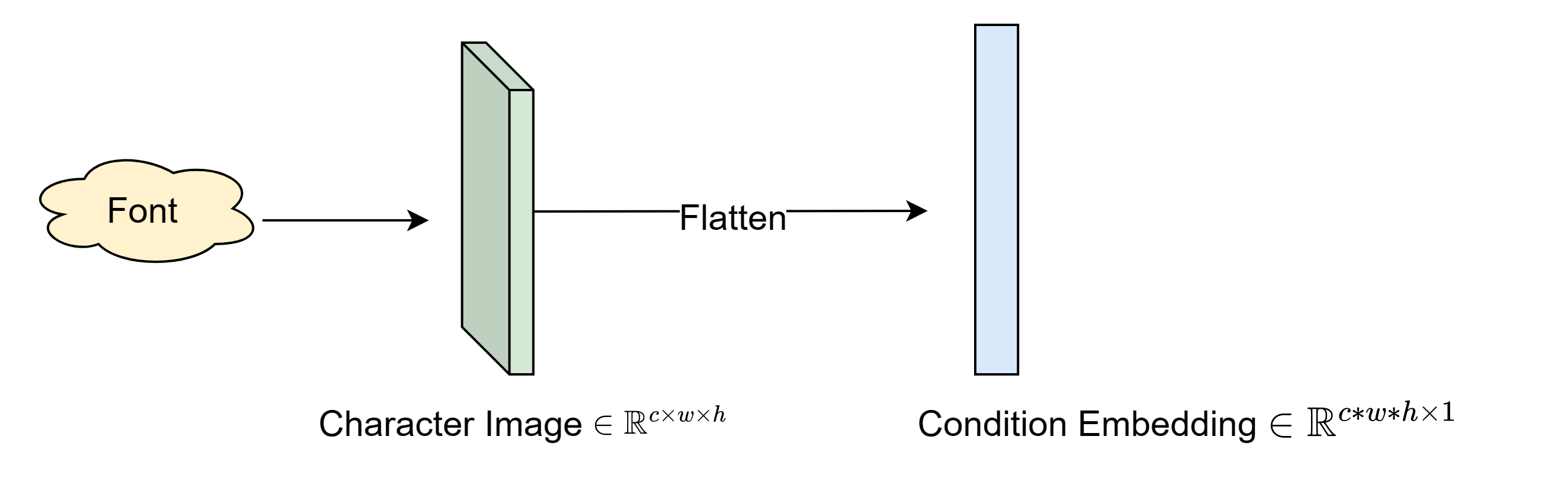}}
\caption{Font image conditioning}
\label{FIC}
\end{center}
\vskip -0.2in
\end{figure}

Given a batch of character label $\in \mathbf{R}^{B}$, we first use a Font that not appear in our training set to generate a character image $\in \mathbf{R}^{C \times W \times H}$ for each label in the batch. Then we flatten the character image and get the pre-embedding $\in \mathbf{R}^{B \times C*W*H}$. In each convolutional step of the Unet, The pre-embedding would be coded by a MLP , boardcasted to the shape of network output on the dimension of channel and added to the output of each convolutional block. This process is illustrated in figure~\ref{FIC}.

\subsubsection{Stroke Information Conditioning}

In our research, we introduce a novel method termed ``Stroke Information Conditioning," which leverages the structural intricacies of Chinese characters for enhancing character generation models. Each Chinese character is composed of strokes, fundamental units that can be categorized into types such as horizontal, vertical, left-falling, right-falling, and others. These types are encoded numerically as 1, 2, 3, 4, and 5, respectively. Consequently, the stroke sequence of any character can be represented as a sequence of these numbers. For instance, the character ``\begin{CJK*}{UTF8}{gbsn} 借 \end{CJK*}" is encoded as ``3212212511" based on its stroke order and types.

In addition to stroke encoding, we also utilize Wubi typing, a method that encodes Chinese characters based on their structure into a sequence of Latin characters. For example, the Wubi code for ``\begin{CJK*}{UTF8}{gbsn} 借 \end{CJK*}" is ``wajg". By combining these two encoding schemes, we obtain a comprehensive representation of each character, such as ``3212212511wajg" for ``\begin{CJK*}{UTF8}{gbsn} 借 \end{CJK*}".

Leveraging this dual encoding system, we train a GPT2-like model to predict the next character in the sequence, thus creating a stroke information embedding. This embedding is designed to capture both the sequential stroke information and the structural characteristics of the characters as per Wubi encoding. The hypothesis is that this combined representation will guide the generative model to produce characters that are structurally and visually accurate at the stroke level. We will take the last hidden state of the last hidden layer, which is a 128-dimension vector, as the condition embedding.
\begin{figure}[H]
    \centering
    \includegraphics[width=1.0\linewidth]{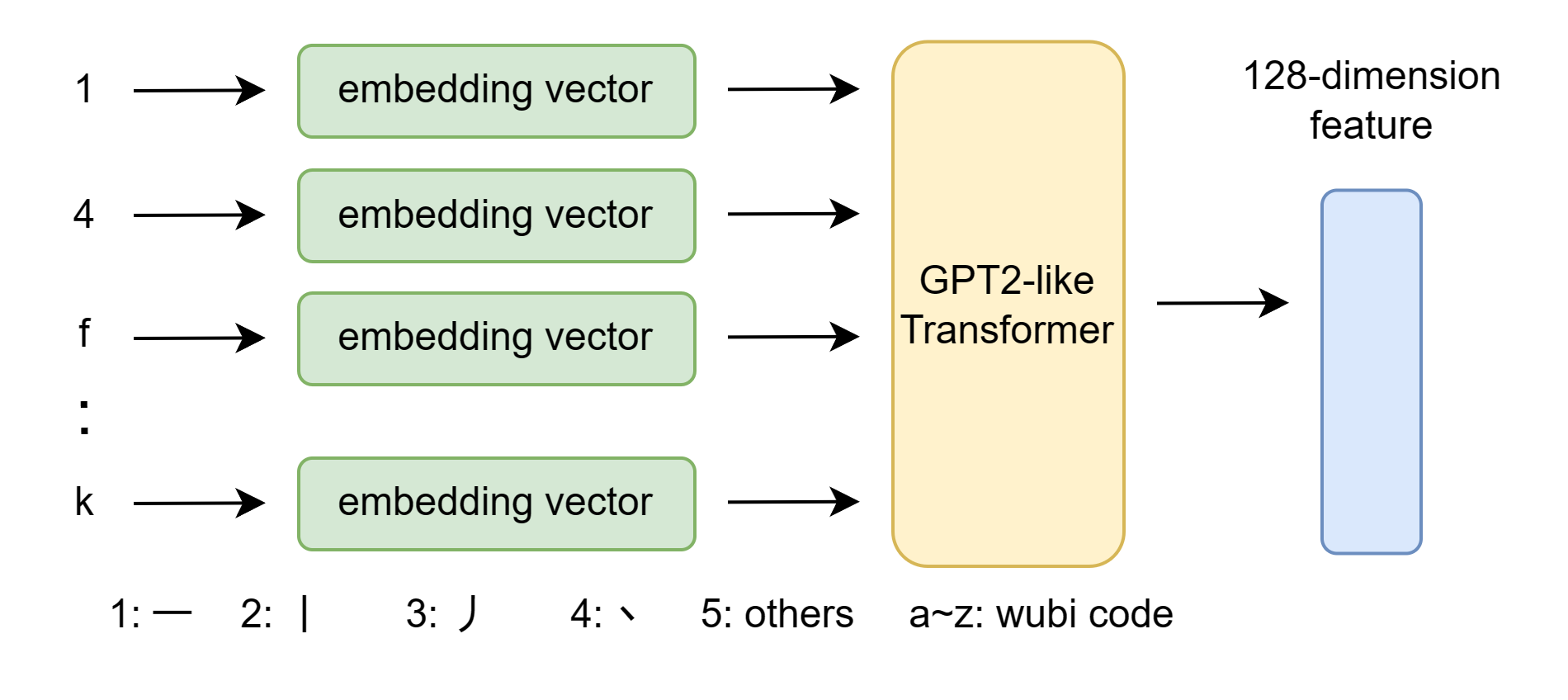}
    \caption{GPT2-like stroke information conditioning}
\end{figure}

\section{Experiments and Results}
\subsection{Datasets}
The pretraining dataset comprises a substantial collection of 196360 images of single characters, spanning three distinct styles: Regular Script, Semi-cursive Script, and Clerical Script. Each character in the dataset is represented with a sample size exceeding 10, ensuring a comprehensive coverage of variations in character design and style. Among these, 163,859 samples originate from internet resources and are attributed to ancient calligraphers, showcasing the authentic and traditional forms of the scripts. The remaining 32501 samples are generated by rendering from various font files.

For the finetuning phase of our model, we utilized a specialized dataset consisting of 196 images. These images are exclusively sourced from the works of President Xu. 

The detailed distributions of our datasets are shown in tabel~\ref{Countofcharacters}

\begin{table}[t]
\caption{Count of characters of each scripts in the pretraining and finetuning dataset. PR stands for pretraining dataset and FI stands for finetuning dataset.}
\label{Countofcharacters}
\vskip 0.15in
\begin{center}
\begin{small}
\begin{sc}
\begin{tabular}{lccccr}
\toprule
            & Regular & Semi-cursive &  Clerical & total \\
\midrule
pr & 77978   & 82285        & 36107     & 196360 \\
fi  & 0       & 107          & 89        & 196\\
\bottomrule
\end{tabular}
\end{sc}
\end{small}
\end{center}
\vskip -0.1in
\end{table}
\subsection{Evaluation}
The evaluation of our model's performance utilizes the Structural Similarity Index (SSIM)\cite{wang2004image}. to assess the similarity between two images. Unlike traditional metrics such as Mean Squared Error (MSE) that solely focus on pixel-level differences, SSIM is designed to capture perceptual changes in structural information, luminance, and contrast. Formally, SSIM is defined as:

\begin{equation}
    SSIM(x, y) = \frac{(2\mu_x \mu_y + c_1)(2\sigma_{xy} + c_2)}{(\mu_x^2 + \mu_y^2 + c_1)(\sigma_x^2 + \sigma_y^2 + c_2)},
\end{equation}
where $x$ and $y$ are the two images being compared, $\mu_x$ and $\mu_y$ are their respective average intensities, $\sigma_x^2$ and $\sigma_y^2$ are their variances, $\sigma_{xy}$ is the covariance, and $c_1$, $c_2$ are constants to stabilize the division with weak denominators.

SSIM ranges from -1 to 1, where a value of 1 indicates perfect similarity. This metric is particularly valuable in our context as it aligns well with the human visual system's quality perception, making it a more appropriate choice for assessing the visual quality of generated images in comparison to other quantitative metrics. SSIM's consideration of perceptual phenomena makes it a robust and reliable metric for evaluating the performance of image generation models, particularly in fields like ours where visual fidelity is paramount.

In evaluating our model's performance, we meticulously partitioned the fine-tuning dataset, which comprises a total of 196 images, into two distinct sets: a training set and a testing set. The training set consists of 157 images, serving as the basis for the model's fine-tuning phase to adapt to the specific stylizations of President Xu's calligraphy. The remaining 39 images constitute the testing set, against which we rigorously compute the SSIM to quantitatively assess the quality of the generated calligraphy.

\begin{figure}[ht]
\vskip 0.2in
\begin{center}
\centerline{\includegraphics[width=0.3\linewidth]{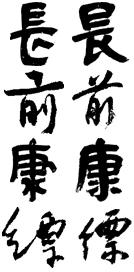}}
\caption{On the left column are the original artworks, while the right column showcases the generated results. None of the characters tested appear in the fine-tuning dataset.}
\label{result}
\end{center}
\vskip -0.2in
\end{figure}

Some generated images are shown in figure~\ref{result} to compare the synthetic character with the genuine artworks from President Xu.

\subsection{Comparison with other methods}
We compare our results to zi2zi \cite{tian2017zi2zi} and CalliGAN \cite{wu2020calligan}, which is shown in table~\ref{comparation}. It is important to note that the results for zi2zi and Calligan reported in our study are directly sourced from the Calligan paper. These results were obtained using their dataset, and not on our dataset.

The results demonstrate that our proposed method achieves comparable performance using a remarkably small dataset, effectively resolving the challenge of data scarcity for the target calligrapher.

\begin{table}[h]
\caption{Comparation with other methods}
\label{comparation}
\vskip 0.15in
\begin{center}
\begin{small}
\begin{sc}
\begin{tabular}{lccr}
\toprule
Method  & SSIM & stylized  trainingset size \\
\midrule
zi2zi & 0.5781   & 5k+       \\
CalliGAN  & 0.6147    & 5k+          \\
proposed+r & 0.4651   & \textbf{157} \\
proposed+t & \textbf{0.4710}   & \textbf{157}\\
proposed+f & 0.4690   & \textbf{157}\\
\bottomrule
\end{tabular}
\end{sc}
\end{small}
\end{center}
\vskip -0.1in
\end{table}

\section{Dataset Collection and Preprocessing}
Besides our proposed method, we put efforts on data collection as well.

We obtained two datasets, a general one for model training and a specific one containing our presidents' handwriting. The general dataset contained 196,360 images including 163,859 of calligraphy works of different artists and a variety of styles such as Regular Script, Semi-cursive Script, and Clerical Script 
The specific dataset of president Xu's artwork was collected through scanning and labeling, with a size of 196 images of characters in total.
\subsection{General Dataset}
We collected online resources of 163,859 calligrahpy artworks. The rest of the dataset consists of fonts converted to images.
\subsection{Specific Dataset}
As we collected the hard copies of the president's calligraphy, we scanned them and obtained images. To extract individual characters for dataset construction, we converted the image into grayscale, applying a binary threshold, which helped separate each item from the background. Then we found contours of each brushstroke and bounded them with smallest rectangles to form initial regions of interest(ROI). This step only captured incomplete characters, so ROIs were further merged based on overlapping areas. By checking if any two ROIs overlapped and comparing to the largest initial ROI, merging overlapped ROIs as long as the area of merged ROI did not exceed the largest initial rectangle, we could extract complete characters as mush as possible. 

\begin{figure}[h]
    \centering
    \includegraphics[width=0.3\linewidth]{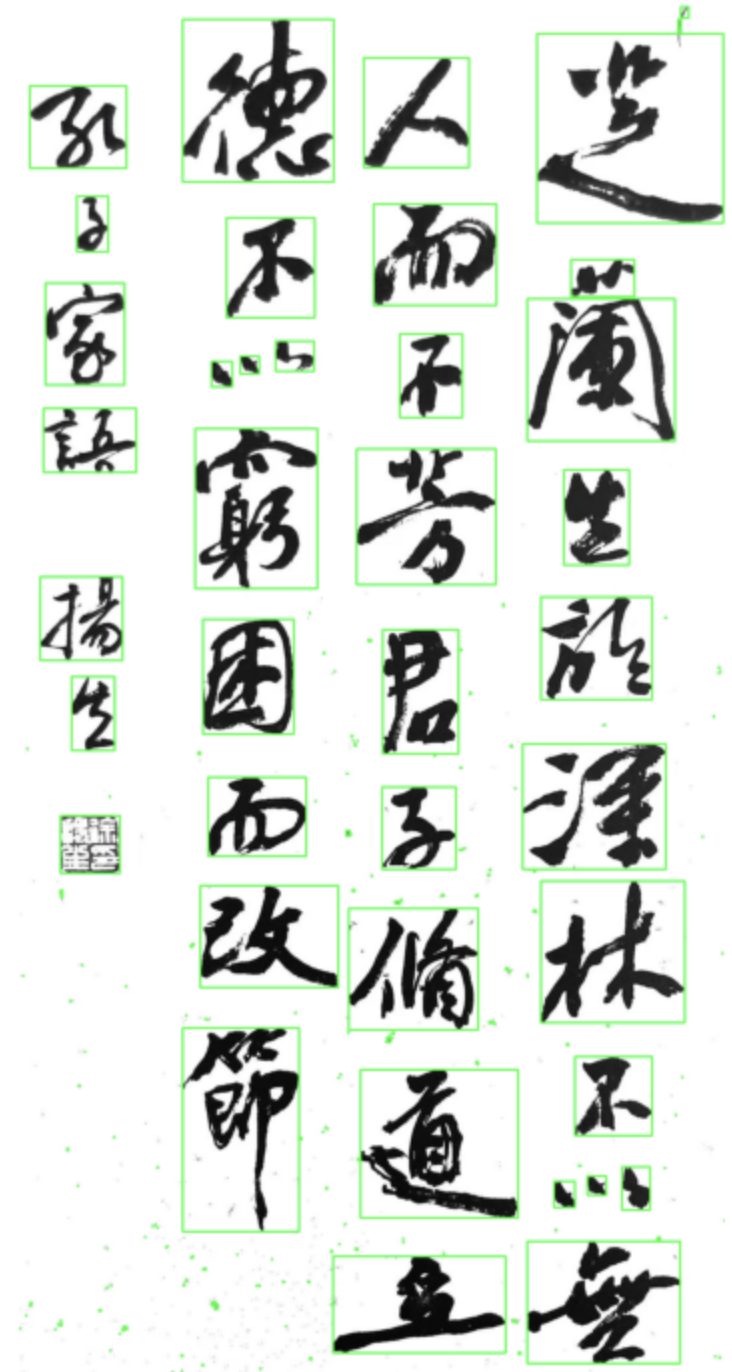}
    \caption{Original Image with ROIs}
\end{figure}
We filtered the extracted characters by size and selected ideal ones, manually labeled them for style transfer.

\section{Conclusion and Future Study}
In this work, we have introduced a novel pretrained diffusion model capable of generating stylized calligraphy with a minimal dataset of approximately 100 labeled artworks. Our method stands out by requiring significantly fewer data compared to existing approaches such as zi2zi and CalliGAN, while still achieving comparable performance. This is particularly notable given the complex nature of calligraphic style transfer and the intricacies involved in capturing the essence of calligraphy with limited samples.

The curated dataset, derived from President Xu's masterpieces, has proven to be highly effective for this task. By leveraging this dataset, our model learns to imbue generated characters with the stylistic nuances that are characteristic of President Xu's calligraphy. Furthermore, the incorporation of stroke information conditioning and font image conditioning has been instrumental in guiding the diffusion process, enhancing the structural correctness of the generated characters, as evidenced by the improved SSIM scores.

Despite these advancements, the quest for perfect structural accuracy in calligraphy generation remains a challenge. While our model marks a significant step forward, there is an evident need for further research to fully resolve the complexities of calligraphic style emulation. Future work will explore more sophisticated conditioning strategies and dataset augmentation techniques to refine the generation process and to push the boundaries of what is possible in the realm of automated calligraphy.

\bibliography{main}
\bibliographystyle{icml2021}

\end{document}